# A Study of Vision based Human Motion Recognition and Analysis

Geetanjali Vinayak Kale, MCOERC, SPPU, Pune, India

Varsha Hemant Patil, MCOERC, SPPU, Pune,, India

## ABSTRACT

Vision based human motion recognition has fascinated many researchers due to its critical challenges and a variety of applications. The applications range from simple gesture recognition to complicated behaviour understanding in surveillance system. This leads to major development in the techniques related to human motion representation and recognition. This paper discusses applications, general framework of human motion recognition, and the details of each of its components. The paper emphasizes on human motion representation and the recognition methods along with their advantages and disadvantages. This study also discusses the selected literature, popular datasets, and concludes with the challenges in the domain along with a future direction. The human motion recognition domain has been active for more than two decades, and has provided a large amount of literature. A bird's eye view for new researchers in the domain is presented in the paper.

## KEYWORDS

## 1. INTRODUCTION

Vision based human motion recognition is a systematic approach to understand and analyse the movement of people in camera captured content. It comprises of fields such as Biomechanics, Machine Vision, Image Processing, Artificial Intelligence and Pattern Recognition. It is an interdisciplinary challenging field having grand applications with social, commercial, and educational benefits. A wide spectrum of applications demands human motion recognition. The applications are spread over domains like sports, medical, surveillance, content based video storage and retrieval, man-machine interfaces, video conferencing, art and entertainment, and robotics (Aggarwal & Nandkumar, 1988; Aggarwal & Cai, 1999; Aggarwal & Ryoo, 2011; Gavrila, 1999; Poppe, 2007; Turaga et.al, 2008). Some of the applications for highlighting the potential impact of human motion recognition are discussed here.

- **Smart Surveillance:** In today's surveillance systems, video contents are viewed continuously by human operators. With the increasing number of cameras, it is impossible for humans to monitor all the contents 24 X 7. Generally, the contents are viewed after a mishap to analyse the event. So, there is an intense requirement of smart surveillance systems from the security agencies. Smart surveillance systems can analyse an event online and provide appropriate intimation using computer based human motion and behavioural analysis. Smart surveillance is required for access control in special areas like military territory, distant human identification, counting the persons and congestion analysis, detection of abnormal behaviour at shopping malls, railway stations, hospitals, government buildings, commercial premises, and schools (Makris & Ellis, 2005; Morris & Trivedi, 2008). Nowadays smart home concept is gaining attention of computer vision community to improve the quality of life of the inhabitant (Guesgen & Marsland, 2016).

- **Behavioural Biometrics:** Nowadays, the use of the gait pattern as a biometric has become popular. The main reason is that the recognition of the gait pattern does not require subject cooperation as compared to the other biometrics (Sarkar et. al 2005).
- **Gesture and Posture Recognition and Analysis:** For a more advanced natural interface with computers and computerized systems, human gesture and posture recognition is an important key. It has promising applications such as gaming, sign language recognition, controlling devices, and others (Ronchetti & Avancini, 2011; Seperi et. al, 2006).
- **Robotics:** Human motion analysis plays an important role in robotics for humanoid robot control, to imitate human motions in a robot in virtual and augmented environments (Hoffman, 2010).
- **Medical:** The medical field uses human motion recognition for the study and analysis of Orthopaedics, Neurology, Musculoskeletal disorders, body posture, and fitness. It is also useful to design intelligent systems to assist elderly people and physically / mentally disabled ones (Najafi et. al, 2003; Lin & Kulic, 2013).
- **Sports and Exercise:** In sports, motion recognition is useful to analyze athletic movements and to design affordable and efficient frameworks for training (Bertini, 2003). An environment for rehabilitation exercise with a feedback system at remote places or in the presence of an expert is designed (Watanabe 2015). Dao (2016) proposed a monitoring system for the exercises of elderly people. These kinds of systems will definitely be useful for patients and old age people.
- **Art and Entertainment:** Motion recognition is useful in analyzing, learning, and an emotional understanding of artistic dance movements as in dances like Bharatnatyam, and Salsa. Kale and Patil (2015) have recognized Bharatnatyam dance sequence from depth data. This also helps to increase the effectiveness of a scene, and the alteration of movements required for quality and the impact of acting.

A large variety of applications have different human motion representation and recognition techniques. Human motion analysis is very general term and application decides the number of body parts involved and duration of movement. Human-computer interaction generally involves only hand gestures, whereas, complicated activity or applications like sports, dance may involve all body parts. Depending on complexity, human motion is conceptually categorized into gestures, actions, activity, interactions, and group activities. Representation and recognition methodologies are decided from tracking and initialization of human body in video. Broad approaches for representation are 2-D Kinematic or stick figure, 3-D kinematic or shape model and image model. Initialization of human using Kinematic method represents human by features like number of joints, its degree of freedom, limb length etc. Whereas, in image model human is represented as image itself and features like shape or region are extracted and stored. Further the recognition can be decided from representation as well as complexity of the motion. Simple actions use sequential or space time single layered approaches. Complicated action requires multi layered approaches. Detailed discussion of recognition methods is given in section 3.4. Even though vision based human motion recognition and analysis has made a much progress, but still it is far away from becoming an off-the-shelf technology. Challenges like occlusion, shadows, lightning, and in-class variance need to be addressed. The remaining paper is organized in to five sections: section 2 discusses the reported literature in the domain, section 3 discusses the human motion recognition framework in detail, a brief discussion on the available datasets is given in section 4, and the paper is concluded with a rigorous discussion in section 5.

## 2. REPORTED LITERATURE

A significant development in human motion recognition and analysis has been seen in the last two decades, and lots of literature in the form of a journal, transaction papers, patents, reviews, and surveys is available. To classify the previous work in the domain, the researchers used criteria like type of models (e.g. stick figure-based, volumetric, statistical), the dimensionality of the tracking space (2-D Vs 3-D) (Gavrila, 1999 & Poppe 2007). Some reviews classify the literature using complexity of the action to be identified (e.g. gesture, action, interaction, group activity) (Aggarwal & Rhoo, 2011). Some survey uses sensor modality (e.g., visible light, infra-red, range), sensor multiplicity (monocular Vs stereo), various applications, number of persons, number of tracked limbs, assumptions (rigid, non-rigid, elastic) for the classification of the available literature (Aggarwal & Cai 1997; Aggarwal & Nandhakumar, 1998; Moeslund, T. B., & Granum,2006; Morris & Trivedi, 2008).

Aggarwal is active in human motion recognition domain since the 1970s, and with a significant amount of work, he has provided time to time reviews of the domain. In his recent review paper with M. S. Rhoo (2011), human motion recognition approaches are classified into two groups: Single layered approaches and Hierarchical approaches.

Cedras and Shah (1995) have described motion based recognition in two steps, the motion information is extraction at first step, and the second step is the matching of an unknown input with the constructed model. Gavrila (1999) has classified the literature into three categories: (a) 2-D approaches without explicit shape models, (b) 2-D approaches with explicit shape models (usually stick figures, wrapped around with ribbons or "blobs") and (c) 3-D approaches (surface-based or volumetric). T. B. Moeslund (2001, 2006) has given a compressive survey of publications in computer vision-based human motion analysis in the form of two survey papers. Taxonomy used in both the survey papers is initialization, tracking, pose estimation, and recognition. The author has classified human motion related applications into surveillance applications (e.g. people counting, congestion analysis), control applications (e.g. Human Computer Interfaces, advanced gaming) and analysis applications (e.g. automatic analysis of orthopaedic patients, and content based retrieval). Wang, Hu, and Tan (2003) proposed a three stage framework for human motion analysis, i.e. human detection (motion segmentation, object classification), human tracking (Model based, Region based, Active contour, Feature based) and human behaviour understanding (Dynamic Time Warping (DTW), HMM, Neural Network, and Semantic description).

Table 1, discusses a few selected papers in the domain. Each paper proposes a new methodology for representation, or it uses a different recognition method. The sample papers for discussion are selected such that they provide a small tour of the domain development for novice users in the domain. First, the authors name and year of publication is shown in column one. Column two shows a pictorial representation of features, motion, or methodology. The next columns discuss the recognition methodology and datasets.

## 3. HUMAN MOTION RECOGNITION

The general framework of human motion recognition systems is discussed in this section. Figure 1(b) shows the steps in human motion recognition systems. For the understanding and analysis of ongoing motion, a scene needs to be captured with an appropriate capturing system. A human needs to be tracked from the scene for further processing. Tracked humans and their motion should be represented efficiently. The represented features are given as an input for motion recognition. The results of human motion recognition highly depend on the selection of an efficient methodology. This selection is decided by parameters like complexity of motion and the extracted features. Figure 1(a) is a pictorial representation of the steps involved in human motion recognition. All the four blocks are discussed in the subsequent section with major emphasis on human motion representation and the recognition methods.

Table 1. Review of selected papers

| Sr. No | First Author Year | Pictorial Representation | Features used & Dataset | Recognition Method | Remarks / Contributions | Limitations |
|---|---|---|---|---|---|---|
| 1 | G. Johansson 1973 | 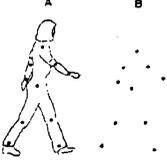 | ● 2-D, Kinematic. LED's are attached on the human body. ● Walking motion is recorded in the dark and played on TV. | - | ● Proves that a human can recognize motion from 2-D motion patterns. ● Pioneer to show joints as a motion recognition feature. | ● Experiment was performed by attaching LED and recording results. ● Not worked on camera captured contents. |
| 2 | Yamato 1992 | 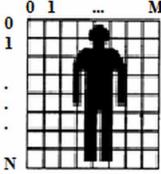 | ● Humanoid Image Model. ● Ratio of the number of black pixels to the number of white pixels in a mesh. ● 5 persons × 6 tennis strokes × 10 times (300 Test data). | Bottom up approach with Discrete HMM. | Recognition rate depends on the training pattern. For same training and testing data, it gives 96.0% results whereas for different training and testing data the results are 70.8%. | ● New model needs to be designed for each considered action ● Dataset considered is too small. ● Experimented on only 2-D images. |
| 3 | D. Marr 1978 Rohr 1994 | 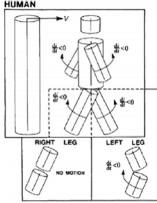 | ● 3-D, Kinematic Hierarchical 3D model based on cylindrical primitives. ● Tested on real as well as synthetic data of a pedestrian for the walking pattern. | Dynamic Time Warping. | ● Proposed representation and recognition of 3-D shapes. ● Used Motion trajectories of body parts for the recognition of motion. | ● Assumes that person walks parallel to image plane ● Person not parallel to image plane and different viewpoint are not considered. |
| 4. | Rao & Shah 2001 | 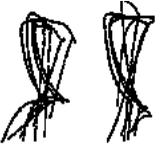 | ● 3-D XYT space time curve. ● 60 different actions by 7 people captured from a different viewpoint. | Single layered, Space time trajectories. Trajectories curvature pattern, and template matching. | ● Main contribution is view invariance recognition. ● Proposed methods didn't show success in more general situations. | ● Methods may fail in general situations ● Experimented on only 7 person trajectories may vary with change in human anthropometry. |
| 5 | Bobick, & Davis 2001 | 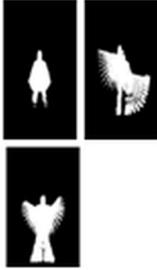 | ● MEI (Motion energy image) and MHI (Motion history image). ● Aerobics steps and kid's rooms. | Single layered, Space time approach, Template matching. | ● Proposed new representation of motion i.e. MHI and MEI. ● Responses of the children appears on the screen. | Approach is very sensitive to action change, also view and size variability. |

**Table 1. Continued**

| Sr. No | First Author Year | Pictorial Representation | Features used & Dataset | Recognition Method | Remarks / Contributions | Limitations |
|---|---|---|---|---|---|---|
| 6 | Laptev and Lindeberg 2003 |  | ● Extracted spatio-temporal interest points are represented on XYT. ● Applied for walking in an outdoor scene. | Single layered space – time local features Euclidean distance between two points in space-time. | Estimated pose of walking people and also detected the motion in the presence of occlusions and dynamic background. | ● Method is variant under Galilean transformation ● Results may vary with change in motion direction. ● Approach may fail for complicated motion. |
| 7 | Shechtman & Irani 2005 |  | ● Spatio temporal patches. ● Tested on video databases with video as a query for walking, with dives in a pool and ballet footage. | Single layered, Space-time-volume correlation. | ● Small video templates compared against large video sequences. ● Variant to large geometric deformation in a video sequence. | Does not handle significant changes in scale and orientation. |
| 8 | Ryoo & Aggarwal 2006 |  | ● Context Free Grammar (CFG) ● Tested for 8 actions approach, depart, hug, punch, kick, and push. Shakehand. | Context free grammar checking. | Experiments show that the system understands continued and recursive composite actions and interactions with a very high recognition rate. | Recognition rate for recursive interaction is still less. |
| 9 | Chaaraoui 2013 |  | ● Fusion of 20 skeleton joints data and silhouette shape divided into S radial bins. ● MSR Action3D 20 different actions performed by 10 subjects up to 3 repetitions. | ● For training Bag of key poses. ● For testing sequence matching using the DTW algorithm. | ● Skeleton approach has shown better results than silhouette. ● Feature fusion has outperformed (AS1-92.38%, AS2- 86.61%, AS3 96.40%). | Proposed approach is sensitive to action class. |

Figure 2 shows a general overview of the components of a human motion recognition system. They include a scene capturing system, human tracking, motion representation methods, motion recognition methods, applications, and datasets. The detailed discussion is given in subsequent section.

**Figure 1. General framework for human motion recognition and its pictorial representation**

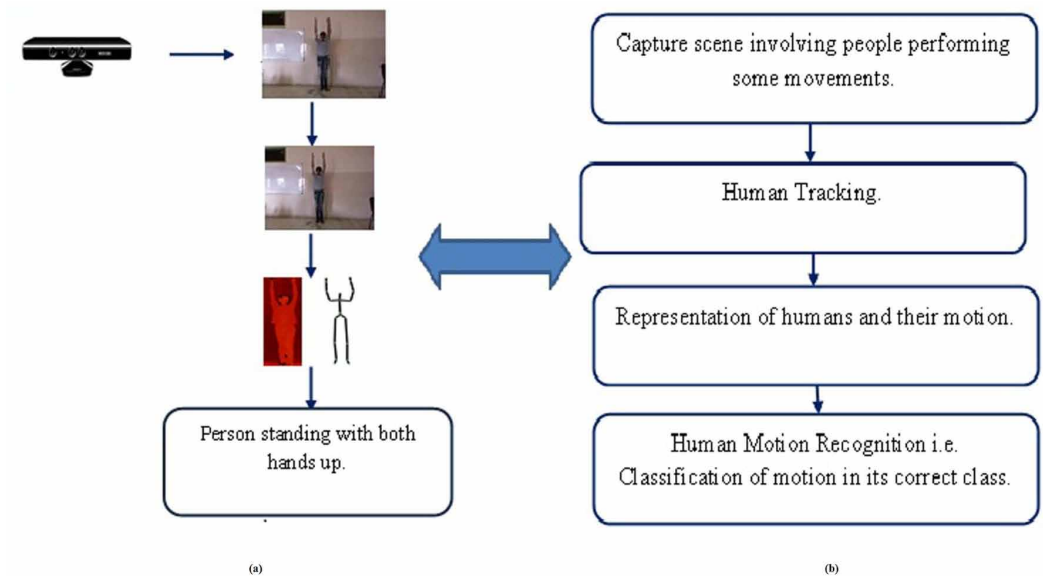

**Figure 2. Overview of human motion recognition and analysis domain**

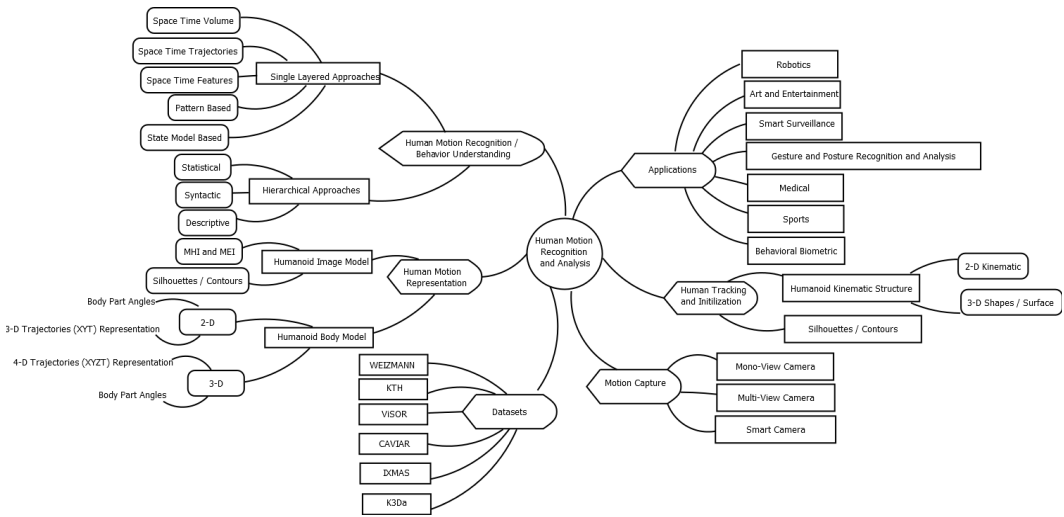

## 3.1. Scene Capturing

The selection of a capturing system depends on the application. A surveillance system application requires more area coverage whereas a device interface requires a more detailed analysis of motion. Nowadays, high resolution cameras are available for surveillance applications. Analysis and controlling applications are either captured by single camera or multiple camera systems. The contents that are captured with a single camera are easy to process, but they may miss the detailed human features. However, the contents that are captured with multiple cameras provide detailed features, but make the systems complicated for processing and analysis. Important initial parameters of human motion analysis from video content is deciding subject as well environmental constraints and camera calibration. Subject constraints can be clothing, known or new subject, subject trained or not trained for performing actions etc. Whereas, environmental constraints include illumination details, constant or variable background. In order to process contents camera details like resolution, frequency of frames, single camera Vs multiple cameras, camera view point etc. needs to be considered. With the advancement in the domain, many sophisticated capture systems like *Microsoft Kinect,* and *ASUS Xtion Pro Live* are available. These kind of systems simplifies initial stages of capturing and human tracking. They provide output in the form of human skeleton joint location details, depth data of objects etc. Researchers can apply representation methodologies and recognition algorithms on obtained data for motion recognition and analysis.

## 3.2. Human Tracking

Human tracking is the identification and initialisation of humans in images or videos. It is challenging, but it is the most important task for the recognition of human motion. Estimation of human body parts from given pose has many challenges like partial occlusion, variations in lighting condition, human physique, missing 3-D information due to 2-D planar display. Still this problem is open challenge for community. Here, we will discuss only the recent developments in the tracking domain. Nowadays, smart cameras are available in the market, which track four to six humans in an indoor scene. Kinect tracks humans in a scene and provides its depth data and skeleton stream as an output for further processing (Han, 2013). The literature has already reported the correctness of depth and the skeleton data of Kinect. Khoshelham, K., and Elberink (2012) has analysed correctness of the depth data and Obdrzalek et. al (2012) analysed accuracy and correctness of Kinect pose estimation. Many researchers have provided human action recognition algorithms using the depth data and skeleton streams obtained from the Kinect sensors (Gao et.al, 2013; Yang & Tian, 2014). The use of Kinect provides assistance for the low level vision problem i.e. scene capture and human tracking. Still Kinect does not provide correct skeleton data for complicated or occluded pose. These kinds of solutions in the form of products will help the researchers to concentrate on high level vision problems.

## 3.3. Human Motion Representation

Human action recognition and representation methods are inter-dependent. The representation methods are broadly categorized in the humanoid body model and the humanoid image model. The humanoid body model uses kinematic structure, and the image model uses humanoid shapes or contour.

### 3.3.1. Humanoid Body Model

The humanoid Body Model uses structural representation and represents a person using his/her joint positions as a set of 2-D (X, Y) or 3-D (X, Y, Z) points in space. The modelling of a stick figure uses human body parts as an estimation methodology. It helps to extract the joint positions of a person from image frame. Kinect sensor version 1 provides (X, Y, Z) points of the human skeleton data for 20 joints, whereas Kinect version 2 provides 25 human joints (Wang, Q, 2015). A different number of joints and their degrees of freedom (DoF) are considered for representation of the human pose. Generally, 20 DoF is used but, for a more detailed analysis like consideration of twisting movement, 34 DoF to 50 DoF models are used. Changes in the person's joint positions while performing an action are stored as space-time trajectories. 3-D XYT or 4-D XYZT space time trajectories are used for the representation of these actions (Rao & Shah, 2001). Johansson is a pioneer in finding components that provide information to humans for understanding human motion. He has experimented it by attaching moving light displays (MLD) to human the body parts (Johansson, 1973). He has recommended that the tracking of joint position are sufficient for humans to distinguish the actions as shown in Figure 3(a).

In several video-based human motion recognition systems the humanoid body model is used to estimation a person's shape. Generally, the model is constructed using shape primitives or surface primitives. Cylinders, cones, ellipsoids, and super quadric are general shape primitives that are used for humanoid shape construction. Polygonal mesh, sub-division surface etc. are surface primitives used for humanoid shape reconstruction. Figure 3(b) shows a humanoid model constructed from cylinders as proposed by Marr and Nishihara (1978). Sminchisescu and Triggs (2003) used a super quadric ellipsoids to represent flesh on skeleton of articulated as shown in Figure 3(c). A major advantage of the humanoid body model representation is that, it is camera view invariant, but robust joint tracking in 3-D is still a topic of research.

Figure 3. (a)-(c) Humanoid Kinematic Structure (Johansson, 1973; Marr & Nishihara, 1978; Sminchisescu & Triggs, 2003) (d) and (e) Humanoid Shape Model(Bobick, 1997; J. Yamato et.al 1992)

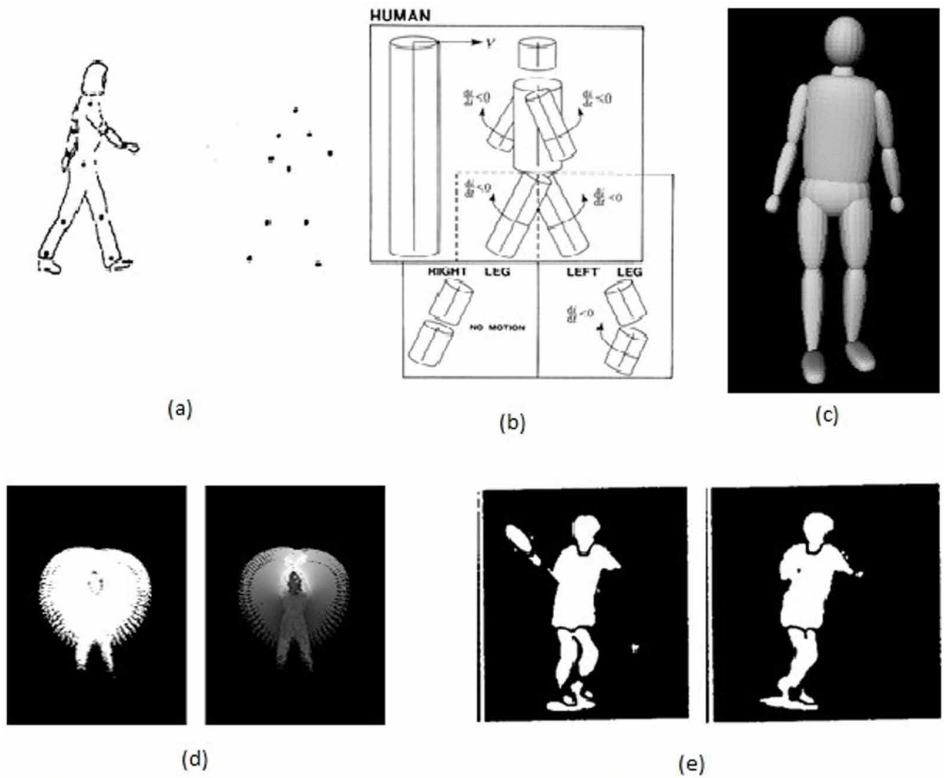

### 3.3.2. Humanoid Image Model

Humanoid image based representation approaches are also known as holistic representation. In these approaches, an action is represented as an image and it does not require detection of an individual body part. Silhouettes or contours of humans performing the action are used for representation. Yamato et.al. (1992) have represented a person using silhouettes as shown in Figure 3(e).

Bobick and Davis (1997, 2001) represented human action using a 2-D motion-energy image (MEI) represented in binary format and a motion history image (MHI). MHI is constructed by projecting a sequence of foreground 2-D images on 3-D space-time volume as shown in Figure 3(d). MHI represents image sequence using decreasing weightage to the sequence of images with less weight to older frames and more weight given to new frames. MEI gives equal weight to all the images in the sequence. A major disadvantage of image based approaches is that they are very sensitive to action change, also view and size variability.

### 3.4. Human Motion Recognition

Human motion can be determined by taking difference between two pixel values in consecutive frames. Kinematic approach represents motion trajectory by 2-D trajectory points $(X, Y, T)$ or 3-D trajectory points $(X, Y, Z, T)$. Each point corresponds to respective joint value in frame for human posture. Image or shape approach represents motion using optical flow or using MHI or MEI. Human motion can be either directly recognized from image sequences, or it can be done in a multiple layer process. Generally, for simple actions, motion is recognized directly from image sequences and they can be viewed as single layer approaches. However, complicated activities can be recognized by using multiple layer recognition methods. Depending on complexity, human motion is conceptually categorized into gestures, actions, activity, interactions, and group activities. Complicated motion can be recognised using multiple layers, by decomposing it into simple actions or gestures. Recognised simple actions or gestures at lower levels are used for the recognition of complicated motions at higher levels. Activities like fighting can be viewed as a sequence of punching and kicking. In the lower layer, the atomic actions of punching and kicking are identified, and then at a higher level an activity is recognized by the sequence of atomic actions.

The recognition methods for simple actions are categorized into space time volume, space time trajectories, space time local features, pattern-based approaches, and state space based approaches. However, for complicated activities and interactions, multi-layer recognition methods are applied. Multi-layer recognition approaches are statistical approaches, syntactic approaches, and descriptive approaches. These approaches are discussed in detail in the subsequent sections.

#### 3.4.1. Trajectory based Recognition

Trajectory based approaches represent human information using the Humanoid Kinematic Structure. The Humanoid Kinematic approach extracts different features like joint angles, limb lengths, and degree of freedom (DoF) from each keyframe. Sheikh et al. (2005) have represented an action using 4-D XYZT space trajectories for 13 joint. XYT trajectories of an action are obtained using affine projection. That in turn helped to find the view-invariant similarity between two sets of trajectories as shown Figure 4(a).

Campbell and Bobick (1995) have represented the human action as curves. Joint positions are tracked using 3-D human body models. For each frame, a body is defined using 3-D estimated models. Static pose in a frame is represented using set of points and action is represented by curves. Each curve is modelled as a cubic polynomial form. The approach is tested on independent pieces of ballet dance. The systems give speed invariant recognition with false acceptance, which is approximately zero.

#### 3.4.2. Space Time Volume

An activity execution in a video can be represented using 3-D space-time volume assembled by concatenating 2-D images with respect to time. There are different methods used by the researchers to store the video in the space time volume fashion. Bobick and Davis have stacked extracted the human silhouettes to track the shape changes as shown in Figure 3(d). They have used template matching approach for recognition of human ativity in kids' room application. Bobick (1997) has further suggested that MEI and MHI are not suitable for complex activities due to overwriting of the motion history. They gives unreliable output for complex activities.

Wu and Shao (2013) have fused the temporally local descriptors called the bag of correlated poses (BoCPs). Gait energy information (GEI) and inversed recording (INV) are added to the MHI. Loss of information in MHI is compensated by GEI and INV. BoCP is created using k-means algorithm. Here, the feature vectors of all the training samples are clustered. The centre of each cluster is defined as a code word and the number of visual code words are the number of the clusters in visual vocabulary. The authors claimed that their approach produces improved results for IXMAS dataset than others.

**Figure 4. Human action recognition methods using (a) 4-D XYTZ approach (Sheikh, 2005) (b) space-time volume correlation (Campbell L. W. and Bobick, 1995), (c) space time feature, (Shechtman and Irani 2005)**

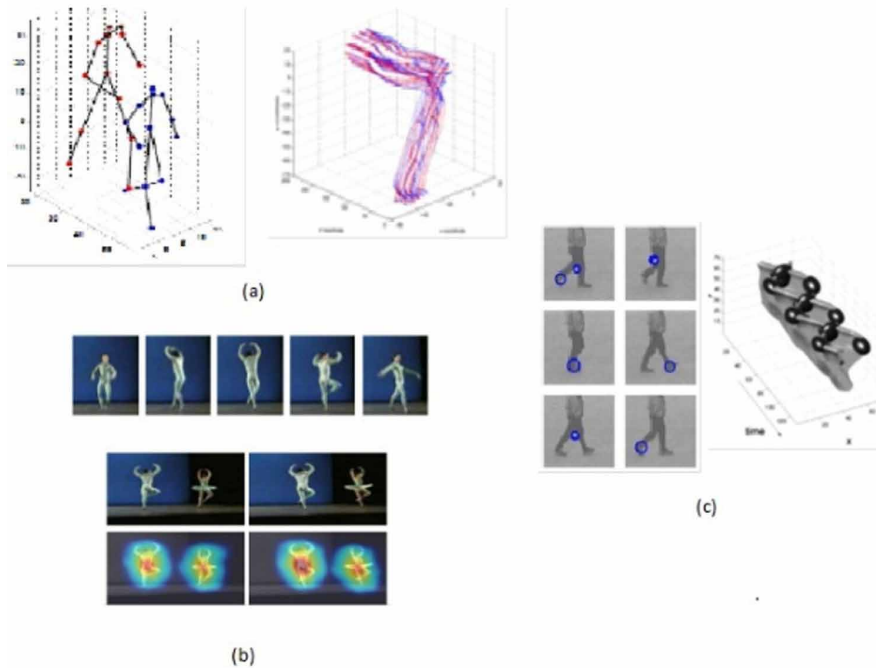

Shechtman and Irani (2005) used a set of 3-D space time volume segments corresponding to a moving human as shown in Figure 4(b). Hierarchical similarity measurement on the space-time volume correlation is used for recognition. The approach is tested on different video databases with the video as the query, like different walking patterns on a beach, dives in a pool, and ballet footage.

### 3.4.3. Space Time Local Features

Local features from each frame are extracted in space time local feature method. They are concatenated for all time to describe the overall motion of human activities. The local features are also referred to as local temporal feature-based approaches. These approaches have shown success in case of change in illumination and presence of noise. Multiple activities are recognised without constructing a human model and subtracting the background.

Laptev and Lindeberg (2004) have applied the space time feature based approach for walking in an outdoor scene. Extracted spatio-temporal interest points are represented on XYT as shown in Figure 4(c). The support Vector Machine (SVM) is used for the classification of motion. More complex activities cannot be modelled using Space-time feature-based approaches

### 3.4.4. Pattern based

Pattern-based approaches for human activity recognition uses whole activity sequence for matching. Sequence of the feature vectors extracted from the query video are compared with templates or sample action executions. Template matching requires major attention towards variability issues. Algorithms like dynamic time wrapping (DTW) can be used for speed invariant action detection. Veeraraghavan et. al (2006) have applied DTW for human motion recognition in human actions with different speeds for activities like throwing an object, picking an object, pushing, and waving. Observed recognition accuracy is very high for all these activities.

### 3.4.5. State Model Based Approaches

State model based approaches represents a human activity using a set of states. The model is statistically trained for corresponding activity class feature vectors. Probabilistic models like Hidden Markov Models (HMMs) and dynamic Bayesian networks (DBNs) are used for recognition.

Yamato et. al (1992) have applied HMM for human motion recognition for the first time. The foreground is converted into meshes and the number of pixels in each mesh is considered as a feature. This approach is applied for the recognition of 6 tennis strokes and 3 persons have performed each tennis action 10 times. Five sequences are used to train HMM and five are used to test the recognition performance.

Lin and Kulic (2014) have segmented on line data and also identified human action. For segmentation the starting and ending point of each action is correctly identified from data. Their approach velocity peaks and zero velocity crossings segmentation of motion. Only the significant DoFs are selected for a given template. Feature-guided Left-right HMM is used in the second phase with joint angles as observation data and hidden states as key poses. The approach is tested for real time interactive feedback in the rehabilitation application for segmentation along with recognition. They have considered rehabilitation movements of 20 healthy persons and 4 rehabilitation patients for test data. Segmentation accuracy achieved is 87% for user specific templates and 79-83% for user-independent templates.

For a complicated activity or interaction, state space based approaches are modelled using multiple layers generally two layers. Simple atomic actions are recognized at the bottom layer from the sequences of the feature vectors. The higher level models treat this sequence of atomic actions as observations, and the maximum likelihood estimation or a maximum posterior probability classifier is constructed to classify the activity. Oliver et. al (2000) have proposed a real-time visual surveillance system for modelling and recognizing human behaviours. Interactions are modelled using both the Coupled Hidden Markov Model (CHMM) and HMM. The complexity of the various interactions decides number of states per chain and number of states in HMM. Here, CHMM shows better results than HMM.

Al Mansure et al. (2013) have applied inverse dynamics to obtain dynamic features for action recognition. These human body features are obtained by using physics-based representation. The dynamic features considered are: the torques obtained from knee and hip joints and gravity, ground reaction forces, and the pose of the remaining body parts. The authors claimed that these features more discriminatively represents human action than the kinematic features. This approach gives good classification results when applied using the HMM classification framework.

### 3.4.6. Hierarchical Syntactic Approaches

Syntactic approaches uses symbols to represent human activity. Each atomic action is represented as symbol and string of symbols represents the action sequence. Researchers generally uses Context-free grammars (CFGs) and stochastic context-free grammars (SCFGs) parsing techniques for the recognition of human activity. Recognition process is designed using more than two layers. Atomic actions are recognized at the bottom layer and are represented as a symbol. At the upper layer, human activities are represented using a set of production rules and are recognized by parsing techniques. These approaches are suitable for a sequential activity, and not for a concurrent activity, but face difficulty when an unknown object appears in the scene.

Ivanov and Bobick (2000), have used the SCFG parsing mechanism for the recognition of multiple interacting objects in surveillance and gesture recognition. HMM is used at low level for detection of atomic feature. The outputs of these detectors provide the input stream for a SCFG parsing mechanism. The authors have handled uncertainty in the input symbol stream by extending stochastic context-free parsing and also handled consistent multi-object interactions.

**Figure 5. Hierarchical approach for human motion recognition - Statistical recognition (Ryoo and Aggarwal 2009)**

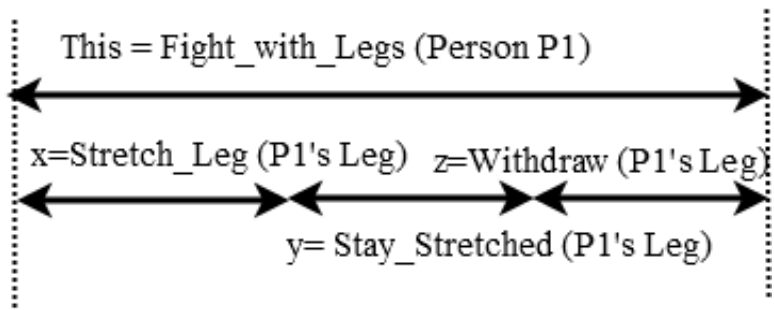

### 3.4.7. Descriptive Approaches

Descriptive approaches represent activities in terms of sub-events /actions with some logical, spatial or temporal relationship. A multi layered design is a must for descriptive approaches. A major advantage of these approaches is that they can handle a concurrent structure.

Ryoo and Aggarwal (2009) have proposed a general framework, for recognition of and human-human interactions and complex human action. The authors used statistical recognition techniques from computer vision and knowledge representation concepts from traditional and artificial intelligence. Figure 5 shows an example of hand shaking. Based on the recognition of gestures at the low level, the high-level of the system hierarchically, recognizes the composite actions and interactions occurring in a sequence of image frames. A major contribution of the authors is the system recognizing human activities including 'fighting' and 'greeting', which are high-level activities. They have a very high recognition rate.

## 4. HUMAN MOTION DATASETS

A standardised dataset is an important need of each domain for the comparison and assessment of an algorithmic performance. With advances in the domain, a variety of datasets are available, some contain an indoor scene, and some are with an outdoor scene. They also differ with motion complexities, number of camera views, moving camera, or a still camera. A detailed analysis of a dataset is not the main focus of the paper. This section is added for the completeness of the paper. Chaquet et. al (2013) provided the detailed analysis of almost 68 different datasets. Here, we have considered only, the mostly referred top few datasets with a variation in the actions for discussion. WEIZMAN (Event and Action), KTH, CAVIAR, ViSOR, IXMAS and K3Da are considered for discussion.

### 4.1. Weizman Datasets

The Weizmann Institute of Science provides two datasets- The Weizmman Event-Based Analysis dataset and Weizmann actions as space-time shapes. The Weizmann event-based analysis dataset was created in 2001. This dataset is used for studying algorithms and segmentation methods.

The Weizmann actions as space-time shapes dataset was created in 2005, to design new algorithms for improving the human motion recognition results. This dataset contains a static background and the moving person's foreground silhouettes. The actions considered and the capturing details are given in Table 2.

### 4.2. KTH Dataset

The KTH Royal Institute of Technology created this dataset in 2004. There are total 25 X 6 X 4 = 600 video files created using a 25 subjects performing 6 actions, and captured in 4 different scenarios.

**Table 2. Human motion recognition Dataset contents and capturing environment details**

| Dataset Name, Year | Type of Actions Considered | Capturing Environment Details |
|---|---|---|
| WEIZMANN Action 2005 | Walking, waving, running, Running in place. | Mono-View, has variability in subject and their clothing. |
| WEIZMANN Event 2001 | walking, jumping, galloping sideways, bending, waving with single and both the hands waving, jumping on place, jumping jack, and skipping, running,. | The view-point is static. Unrealistic action analysis. |
| KTH 2004 | Walking, jogging, running, boxing, hand waving, and hand clapping. | Mono-View, Non-realistic, Captured in controlled environment. Added complexity by changing lighting conditions and clothing of subjects, unrealistic action analysis. |
| CAVIAR -Context Aware Vision using Image-based Active Recognition 2007 | Meeting, Walking, entering, in shop, browsing, slump, left object alone, fighting, window shop, and exiting from shop. | Captured at the entrance lobby of the INRIA Labs at Grenoble, France and a shopping centre in Lisbon. Realistic action analysis. |
| ViSOR -Video Surveillance online repository 2005 | Video sequences taken from a real surveillance setup, University of Modena and Reggio Emilia campus monitoring. | Multi camera systems composed by 8 different surveillance cameras with a non-overlapped field of views. |
| IXMAS 2006 | Scratching head, sitting down, getting up, throwing, turning around, walking, waving, punching, kicking, pointing, picking up,, nothing, checking wrist watch, crossing arms. | Multi-view, captured using 5 cameras, controlled conditions (indoor), Realistic. |
| Kinect 3D Active (K3DA) 2015 | Participants performed standardized tests, including the Short Physical Performance, Timed-Up-And-Go, vertical jump and other balance assessments, which were recorded using depth sensor technology. | Motions collected from young and older men and women ranging in age from 18 - 81 years. |

The sequences were taken at 25 frame rate using static camera and same backgound. The details of actions considered, and the capturing environment are shown in Table 2.

## 4.3. CAVIAR Dataset

(CAVIAR) Context Aware Vision using Image-based Active Recognition was the first dataset recorded in more complex environments. The CAVIAR dataset includes 9 activities. The videos are recorded at two different places. The first at the entrance lobby of the INRIA Labs at Grenoble France and the second part is recorded at a shopping centre in Lisbon. The CAVIAR project has produced huge publications that focuses on variety of applications, such as target detectors, activity recognition, monitoring of human activity, segmentation, and tracking of movement, or multi-agent activity recognition.

## 4.4. VISOR Dataset

ViSOR (Video Surveillance online repository) dataset is provided by the Image Lab of the University of Modena and Reggio Emilia. It is more realistic and contains videos captured with surveillance cameras in indoor and outdoor scenes. The researchers have used these datasets for the study of human behaviour analysis, detection of smoke, human tracking, event analysis, counting people, pedestrian crossing, and identification of human, also human action recognition.

## 4.5. IXMAS Dataset

IXMAS is created for view-invariant human action recognition. Ground truth, and silhouettes are provided in a BMP format (390 X 291) and reconstructed volumes (64 X 64 X 64) are provided in the MATLAB format. This dataset contains 13 day today activities performed by 11 subjects 3 times each. The actors freely choose position and orientation. The details of the actions considered and the capturing environment are as shown in Table 2.

## 4.6. Kinect 3D Active (K3DA)

Kinect 3D Active considers the clinically-relevant motions dataset, which is prepared using the Microsoft Kinect One sensor. Dataset is released to the community as an open source solution for benchmarking detection, quantification and recognition algorithms. The dataset includes motions collected from young and older men and women ranging in age from 18 - 81 years. Participants performed actions like Timed-Up-And-Go, vertical jump and other balance assessments which were recorded using depth sensor technology and extracted to generate motion capture data, sampled at 30 frames-per-second.

## 5. DISCUSSION

Human motion recognition from video content is challenging domain with potential applications. There is the need to cope with the challenges in segmentation, modelling, and occlusion handling. The system requires robust solutions, which can help in product design. Some of the challenges in the domain are discussed in the following section, and it will provide a future direction for novice researchers in the domain.

## 5.1. Designing Invariant System

Designing an invariant system is one of the major challenges as same action class shows the wide variability in the features. From the reported literature, in-class variability occurs due to three main factors: (a) Action execution rate, (b) Human anthropometry, and (c) Camera view point. The humanoid image model representations are highly sensitive to in-class variability than the humanoid body model.

### 5.1.1. Action Execution Rate

Action execution rate for a particular action may vary with time for the same person, and there can be a difference in the action execution rate of different persons. DTW is applied by many researchers to overcome variability due to the execution rate. Pham et. al (2014) applied DTW and voting algorithms on the 3D human skeletal models. They claimed that their methods have outperformed the previous methods in both recognition accuracy, and computational complexity.

Variability in the execution rate may appear due to an expert level in performing an action, or it may vary with the situation. Dance steps or exercises in rehabilitation may show a significant difference in the execution rate due to an expert level in a performing an action. Whereas, in a running action, situations like a street dog running behind, compels a person to run faster.

Even though these approaches are successful in lab experiments, they may fail for real world practical problems.

### 5.1.2. Human Anthropometry

Anthropometric (body measurement) invariance is one of the important parameters to be considered for a robust system design. Anthropometric variances are seen due to change in size, shape, gender, and other similar parameters. In applications like surveillance anthropometry, it may not be very significant for action recognition, but it plays a major role in motion analysis applications. This problem has received meager attention of the researchers.

### 5.1.3. Camera View Point

Camera view point plays a very important role for applications like surveillance. Successful approaches for a single view point miserably, fail for other view-points. Some researchers have proposed systems for view-invariant recognition of motion. Rao and Shah (2001) designed the view invariant representation and recognition method using the spatiotemporal curvature of motion trajectories. The incremental learning approach is used for training the system for a given action captured from the different viewpoints for different people. But, the proposed methods didn't show any success in more general situations.

## 5.2. Intention Reasoning

Intention reasoning plays a very important role in security applications. For most of the cases, the system is trained for specific actions, and may fail for different unpredicted actions. For the detection of fighting, if the system is trained for kicking and punching, it may fail to detect hitting with an object in fighting even though it is a part of fighting. Today's systems may fail to identify the difference between Karate practice and actual fighting due to ignorance of the intention reasoning parameter.

So, there is a strong need of designing robust generalised systems with intention reasoning.

## 5.3. Providing Product For Real World Problems

Various available algorithms for human motion representation, and recognition are mainly, driven by specific applications or datasets. Many researchers and organisations are actively involved in the domain, and have provided a variety of datasets. Chaquet et.al (2013) have reported almost 68 datasets with a variety of motions including complex behavior. In order to design and deploy vision based systems for various surveillance, and control and analysis applications in real time environment, there is the need of a rigorously prepared, standardised common dataset to assess and compare the algorithmic performances. Advances in other domain can be applied for more accurate results. Saba et. al (2016) have applied online accurate feature classification paradigm for automated stratification of liver diseases in ultrasound. This will lead the Computer Vision community to provide a robust solution for real world problems in various applications.

## 6. CONCLUSION

The recognition of human action from video content is a significant area of research in computer vision, and has shown considerable progress in the domain. The paper discussed different representation and recognition strategies classified according to the complexity of action. Hierarchical approaches have shown great success for the recognition of complicated actions and interactions. Techniques like bag-of-words, and HMM that have shown success in speech and text recognition are successfully applied for action recognition. Advances in fields like Artificial Intelligence and Machine Learning needs to be applied for human motion recognition. Now, with efforts of research community in human motion recognition, products with intelligent camera systems for social and commercial applications should be made available in market.

## ACKNOWLEDGMENT

This work is partially funded by the Board of College and University Development, SP Pune University, Pune, Maharashtra, India. Grant Number is 14ENG002310.

*Geetanjali Vinayak Kale has completed her ME (CE) in 2004. Currently she is research scholar of Savitribai Phule Pune University, Pune and working as Assistant Professor at Computer Engg. Department of PICT. Her area of interest is Computer Vision, Machine Learning, and Computer Graphics.*

*Varsha Hemant Patil completed her PhD in 2009. She is currently working as Vice Principal and head of Computer Engg. MCOERC, Nashik. Currently, she is also Chairman Board of Studies Computer Engg. Savitribai Phule Pune University, Pune.*